\documentclass[11pt]{article}

\usepackage[preprint]{acl}

\usepackage{times}
\usepackage{latexsym}

\usepackage[T1]{fontenc}
\usepackage[utf8]{inputenc}

\usepackage{microtype}
\usepackage{inconsolata}

\usepackage{graphicx}
\usepackage{amsmath,amssymb}
\usepackage{booktabs}
\usepackage{multirow}
\usepackage{url}

\title{ROG: Retrieval-Augmented LLM Reasoning for Complex First-Order Queries over Knowledge Graphs}

\author{Ziyan Zhang\textsuperscript{1}, Chao Wang\textsuperscript{2}, Zhuo Chen\textsuperscript{2}, Chiyi Li\textsuperscript{2}, Kai Song\textsuperscript{1} \\
$^1$School of Information Science and Engineering, Chongqing Jiaotong University\\ $^2$State Grid Chongqing Electric Power Company\\
}

\begin{document}
\maketitle

\begin{abstract}
Answering first-order logic (FOL) queries over incomplete knowledge graphs (KGs) is difficult, especially for complex query structures that compose projection, intersection, union, and negation.
We propose ROG, a retrieval-augmented framework that combines query-aware neighborhood retrieval with large language model (LLM) chain-of-thought reasoning.
ROG decomposes a multi-operator query into a sequence of single-operator sub-queries and grounds each step in compact, query-relevant neighborhood evidence.
Intermediate answer sets are cached and reused across steps, improving consistency on deep reasoning chains.
This design reduces compounding errors and yields more robust inference on complex and negation-heavy queries.
Overall, ROG provides a practical alternative to embedding-based logical reasoning by replacing learned operators with retrieval-grounded, step-wise inference.
Experiments on standard KG reasoning benchmarks show consistent gains over strong embedding-based baselines, with the largest improvements on high-complexity and negation-heavy query types.
\end{abstract}

\section{Introduction}
Knowledge graphs (KGs) represent facts as relational triples and support structured reasoning over large knowledge bases such as Freebase and WordNet \citep{bollacker2008freebase,miller1995wordnet}.
In many real deployments, KGs are large, sparse, and incomplete, which makes it difficult to answer queries that require multi-hop reasoning.
Answering first-order logic (FOL) queries \citep{enderton2001mathematical} on incomplete KGs is therefore challenging, especially for complex operator compositions \citep{liang2024kg}.
Most existing approaches rely on geometric embeddings \citep{hamilton2018embedding}, but performance and generalization often degrade on complex queries and heterogeneous graphs \citep{choudhary2023complex,choudhary2022probabilistic,teru2020inductive}.

\begin{figure*}[t]
  \centering
  \includegraphics[width=\linewidth]{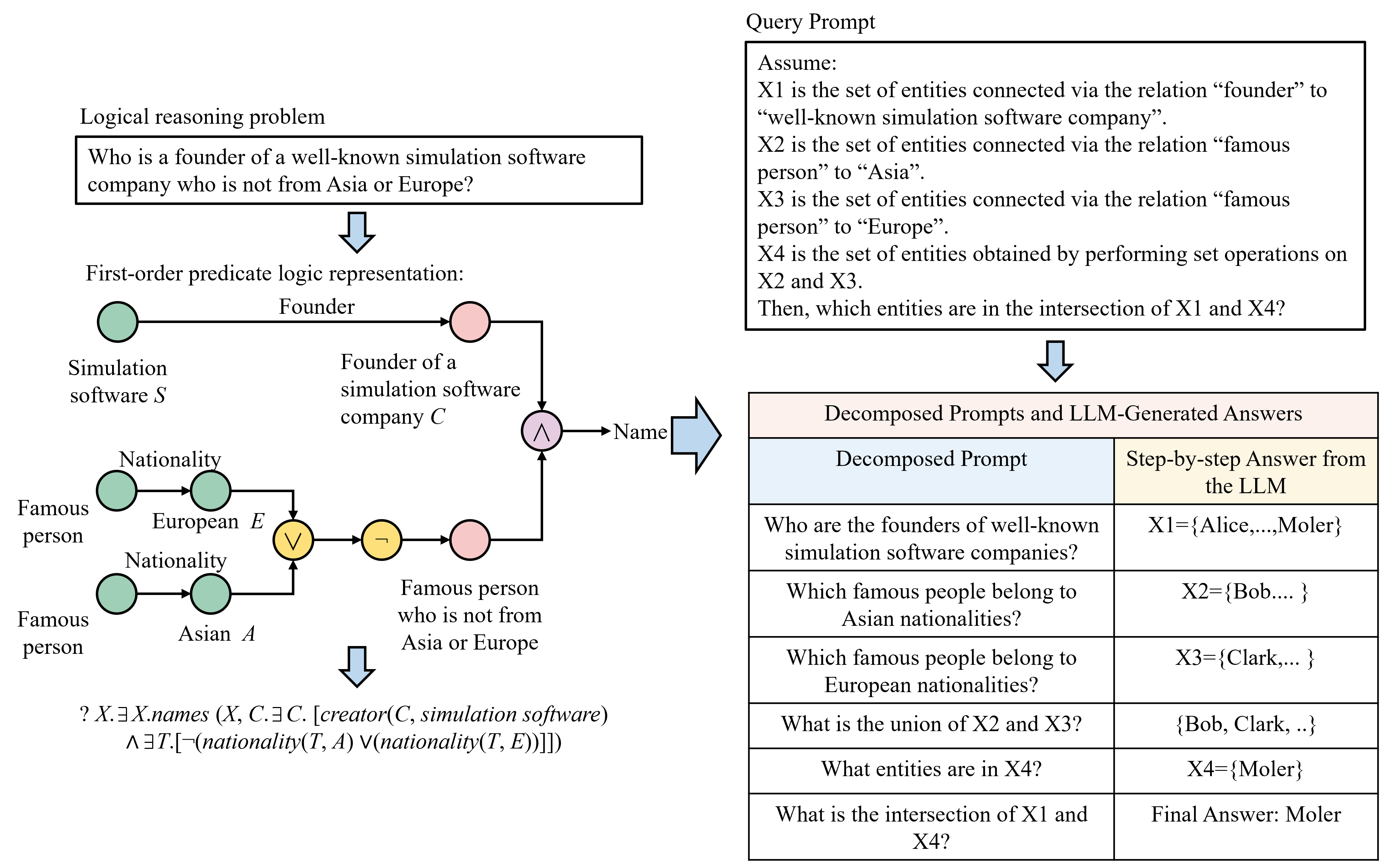}
  \caption{Workflow of Logical Reasoning Decomposition and LLM-Based Answer Generation.}
  \label{fig:fig1}
\end{figure*}

We propose \textbf{ROG}, which combines (i) query-aware neighborhood retrieval and (ii) LLM-based chain-of-thought inference over retrieved evidence.
ROG abstracts entity/relation names to identifiers, decomposes a multi-operator query into single-operator sub-queries, and answers them sequentially with cached intermediate results, guided by prompting strategies \citep{huang2024trustworthy}.
This decomposition helps the model focus on executing a single logical operation.

At a high level, ROG treats logical query answering as a sequence of small, verifiable set operations.
Each sub-query is answered using only a compact, query-relevant neighborhood, and the resulting intermediate answer sets are cached and reused.
This design keeps the LLM grounded in retrieved evidence while still exploiting its flexible reasoning ability.

Unlike purely embedding-based approaches, ROG does not require learning task-specific operator parameterizations for each dataset.
Instead, it delegates the combinatorial composition of operators to decomposition plus prompting, and uses retrieval to supply the minimal evidence needed for each step.
As a result, the framework is easy to adapt to different KGs and query distributions, as long as neighborhoods can be retrieved and serialized consistently.
In addition, the same prompting and execution strategy can be applied across operator types, simplifying system engineering.

Our contributions are:
\begin{itemize}
  \item An LLM-driven KG reasoning framework for FOL query answering that combines decomposition, retrieval, and step-by-step inference.
  \item A practical abstraction strategy that replaces entity/relation names with identifiers to reduce hallucination and improve cross-KG robustness.
  \item Empirical results on standard benchmarks showing improved MRR over strong embedding baselines, especially on high-complexity query types.
\end{itemize}

\section{Related Work}

\paragraph{LLM prompting and decomposition.}
LLMs can solve many NLP tasks via prompting \citep{wu2024power}, and chain-of-thought prompting improves multi-step reasoning \citep{wei2022chain}.
Recent work explores decomposition and tool/symbolic augmentation \citep{creswell2022selection, gao2023pal, yang2024harnessing, yang2024can, xiong2025deliberate, zhou2022least, khot2022decomposed, wang2022self, yao2023tree}.
ROG follows this line but targets structured KG reasoning: we explicitly decompose logical operators, ground each step in retrieved triples, and cache intermediate sets to stabilize multi-step inference.

\paragraph{Logical KG reasoning.}
Embedding-based methods model projection and set operations over KGs \citep{nickel2011rescal, bordes2013translating, nickel2017poincare, ren2020query2box, ren2020beta, minervini2021complex}.
These approaches are typically efficient at inference time but depend on learning operator behaviors from data and can be sensitive to changes in graph schema or query distributions.
Hybrid approaches incorporate KG neighborhoods for reasoning \citep{yasunaga2021qagnn}, and extensions address temporal KGs \citep{xiongtilp, xiong2024teilp, xiong2024large}.
In contrast, ROG performs retrieval-augmented LLM inference directly over a compact, query-relevant subgraph.
This shifts complexity from training to query-time reasoning, which can be advantageous when deploying across multiple KGs.

\begin{figure*}[t]
  \centering
  \includegraphics[width=\linewidth]{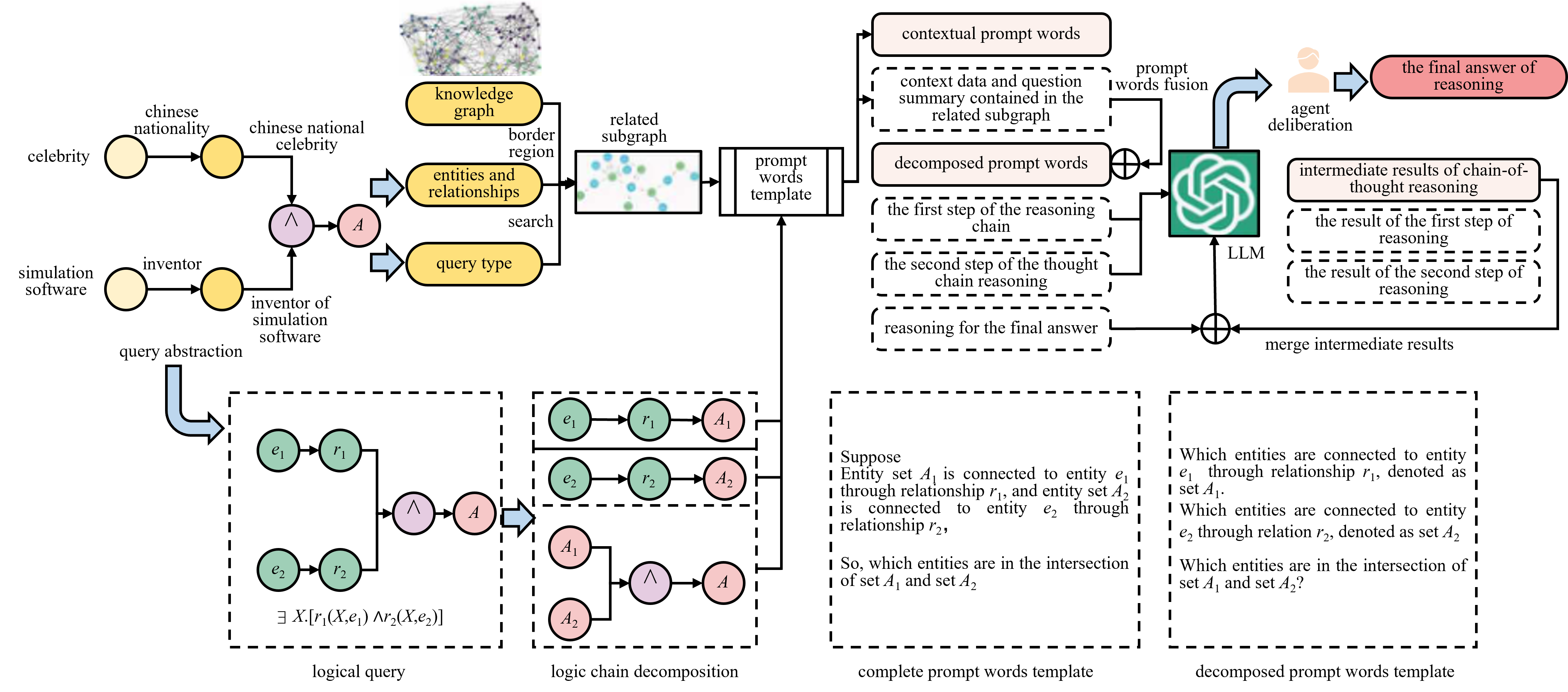}
  \caption{Data flow of the ROG model.}
  \label{fig:fig2}
\end{figure*}

\section{Methodology}

ROG answers FOL queries over a KG $G \subseteq E \times R \times E$ using four operators: projection ($P$), intersection ($\wedge$), union ($\vee$), and negation ($\neg$).
Instead of reasoning over the full KG, ROG retrieves a compact query-relevant neighborhood and performs retrieval-augmented inference with an LLM.

\subsection{Problem Setup and Operators}
A query is defined by a small set of grounded entities and relations and returns an entity set that satisfies the logical constraints.
We focus on standard query families composed of $P$, $\wedge$, $\vee$, and $\neg$ operators.
Intuitively, projection follows a relation to obtain a candidate set, intersection/union combine candidate sets, and negation removes entities that satisfy a constraint.
In practice, complex queries are built by composing these primitives into deeper reasoning chains.

Operationally, each intermediate set can be viewed as a candidate pool of entity identifiers.
ROG treats the execution of each operator as producing or transforming such pools, which makes intermediate states explicit and reusable.
This aligns naturally with retrieval: for a given operator application, we only need evidence that is local to the current pool and the queried relations.

\subsection{Abstraction}
We replace concrete entity and relation surface forms with unique identifiers.
This forces the LLM to reason from the provided evidence rather than memorized world knowledge, reducing hallucination and improving portability across heterogeneous KGs.
Abstraction also makes prompts more uniform across datasets, which reduces the amount of prompt engineering needed when changing KGs.

\subsection{Neighborhood Retrieval and Query Decomposition}
Given a query $q$, we retrieve a query-relevant neighborhood around the entities and relations appearing in $q$.
To control context length, ROG decomposes a complex multi-operator query into a chain of single-operator sub-queries (e.g., $3p$ becomes three consecutive $1p$ queries; a $3i$ query becomes three projections followed by one intersection).
Each sub-query is answered with the evidence restricted to its local neighborhood, and intermediate answer sets are cached and injected into later prompts.
Decomposition ensures that each prompt is short and operator-focused, which improves reliability.

\paragraph{Retrieval granularity.}
In practice, we retrieve a small $k$-hop neighborhood for each grounded entity involved in the current sub-query and optionally expand around newly produced intermediate entities.
We cap the number of retrieved triples to fit the LLM context window and prioritize edges that match the relations appearing in the query.
This makes the evidence both compact and query-aware.
When the neighborhood is large, truncation prevents the LLM from being distracted by unrelated facts.

\paragraph{Deterministic execution order.}
Decomposition follows the syntactic structure of the logical query and produces a fixed execution plan.
For composite queries, we compute projections first to obtain candidate sets and then apply set operations (intersection/union) on the cached intermediate sets.
For negation, we explicitly represent the ``to-be-removed'' set and apply subtraction at the end of the corresponding sub-query.
A deterministic plan also simplifies caching because intermediate results can be indexed by step type and arguments.

\begin{table*}[t]
\centering
\small
\begin{tabular}{llccccccccc}
\toprule
Dataset & Model & 1p & 2p & 3p & 2i & 3i & ip & pi & 2u & up \\
\midrule
\multirow{4}{*}{FB15k}
& GQE & 57.2 & 55.9 & 29.9 & 52.4 & 47.7 & 41.8 & 44.5 & 29.8 & 31.2 \\
& Q2B & 58.1 & 48.7 & 43.3 & 65.1 & 56.9 & 47.1 & 50.2 & 35.2 & 29.9 \\
& CQD & 72.4 & 56.9 & 41.5 & 64.6 & 61.3 & 54.4 & 58.8 & 55.7 & 36.9 \\
& ROG & \textbf{81.4} & \textbf{67.7} & \textbf{49.2} & \textbf{75.6} & \textbf{72.3} & \textbf{62.0} & \textbf{65.1} & \textbf{69.4} & \textbf{45.6} \\
\midrule
\multirow{4}{*}{NELL995}
& GQE & 52.5 & 29.5 & 15.3 & 35.6 & 38.8 & 28.5 & 25.4 & 26.9 & 18.6 \\
& Q2B & 53.1 & 31.0 & 22.4 & 51.4 & 46.9 & 31.1 & 29.1 & 37.6 & 21.5 \\
& CQD & 66.4 & 32.7 & 26.8 & 55.8 & 51.2 & 37.7 & 33.6 & 45.4 & 39.2 \\
& ROG & \textbf{83.3} & \textbf{59.2} & \textbf{41.5} & \textbf{61.1} & \textbf{58.7} & \textbf{42.9} & \textbf{50.1} & \textbf{62.2} & \textbf{57.2} \\
\bottomrule
\end{tabular}
\caption{MRR comparison between ROG and traditional methods on nine typical query types.}
\label{tab:typical}
\end{table*}

\subsection{Chain-of-Thought Prompting}
After decomposing a query into a chain of simple sub-queries, ROG performs sequential inference with the LLM.
We use prompt templates to (i) serialize the retrieved neighborhood into a compact, deterministic text format and (ii) present a single logical operation to be executed at each step.

\paragraph{Evidence serialization.}
The retrieved neighborhood is rendered as a list of triples and small adjacency lists, using the abstract identifiers from the previous stage.
This keeps the context structured and makes it easy to verify whether an output entity is supported by the evidence.
It also allows downstream components to audit the reasoning by checking whether each returned entity appears in the evidence.

\paragraph{Step-wise reasoning with caching.}
For each sub-query, the prompt includes: the operator type, its arguments (entity IDs, relation IDs, and/or previously computed answer sets), and the local evidence.
The LLM returns an intermediate answer set, which is stored in a cache.
Later prompts reference cached sets using placeholders, so that multi-step queries remain consistent and do not require the model to ``re-derive'' earlier answers.
Caching also reduces redundant computation when a query contains repeated sub-structures.

\paragraph{Multi-agent consensus (optional).}
For domain-specific scenarios, ROG can instantiate multiple agents (each pairing an LLM with the same evidence) and aggregate their outputs via a simple consensus rule (e.g., majority voting over candidate entities).
This can improve robustness when the base model is sensitive to prompt phrasing.
In addition, disagreement among agents can be used as a signal that the retrieved evidence is insufficient or ambiguous.

\paragraph{Output constraints.}
To avoid verbose generations and reduce post-processing ambiguity, we constrain the LLM output to a simple list format.
When a sub-query expects a set of entities, the model outputs only entity identifiers (one per line).
When a sub-query expects an empty set, the model outputs an explicit \texttt{NONE} token.
This lightweight constraint improves determinism and makes cached intermediate sets easy to parse and reuse.
It also makes evaluation straightforward because outputs can be directly compared as sets.

\section{Experiments}

\subsection{Experimental Setup}
We evaluate ROG on FB15k \citep{bordes2013translating} and NELL995 \citep{xiong2017deeppath}, following the standard benchmark protocols for complex logical query answering.
We use ChatGLM \citep{glm2024chatglm} as the base LLM and store KGs in Neo4j for efficient neighborhood retrieval.
All experiments are run on a workstation with an NVIDIA GeForce RTX 4090 GPU.

\paragraph{Baselines.}
We compare against representative embedding-based methods, including GQE, Query2Box (Q2B), and CQD.
These baselines cover both neural set-operator designs and decomposition-style embedding methods.

\paragraph{Query types and evaluation.}
We evaluate typical projection, intersection, union, and composite query types (e.g., $1p$--$3p$, $2i$--$3i$, $2u$, $ip$, $pi$), and additionally report results on negation-heavy query types.
We use Mean Reciprocal Rank (MRR) as the main metric.
MRR is sensitive to the rank position of the correct entity and is widely used for KG query answering evaluation.

\paragraph{Implementation details.}
For retrieval, we use Neo4j to execute neighborhood expansion queries and return a bounded set of triples for each sub-query.
For prompting, we use a fixed template that includes: (i) a short instruction describing the operator to apply, (ii) the serialized neighborhood evidence, and (iii) the current arguments (entity/relation IDs or cached intermediate sets).
To improve stability, we constrain the output format to a newline-separated list of entity identifiers.
We also keep the prompt structure fixed across datasets to avoid tuning to a specific benchmark.

\paragraph{Practical considerations.}
We found that prompt stability depends on keeping evidence compact and avoiding distractor relations.
Therefore, retrieval prioritizes triples whose relations occur in the current sub-query and truncates long adjacency lists.
When the retrieved evidence is insufficient (e.g., because the KG is incomplete or the neighborhood cutoff is too strict), ROG may under-generate candidate entities; this behavior is preferable to hallucinating unsupported entities.
In addition, the deterministic execution plan ensures that the same sub-query always triggers the same retrieval strategy.

\begin{table}[t]
\centering
\small
\begin{tabular}{llccccc}
\toprule
Dataset & Model & 2in & 3in & inp & pin & pni \\
\midrule
\multirow{2}{*}{FB15k}
& BetaE & 29.8 & 21.7 & 19.1 & 14.4 & 18.4 \\
& ROG   & \textbf{36.4} & \textbf{35.7} & \textbf{22.3} & \textbf{19.6} & \textbf{25.0} \\
\midrule
\multirow{2}{*}{NELL995}
& BetaE & 29.6 & 27.9 & 23.4 & 21.1 & 15.6 \\
& ROG   & \textbf{34.7} & \textbf{32.8} & \textbf{29.5} & \textbf{27.3} & \textbf{19.5} \\
\bottomrule
\end{tabular}
\caption{MRR comparison between ROG and BetaE on complex query types.}
\label{tab:complex}
\end{table}

\subsection{Effectiveness of ROG for Logical Reasoning}

We report Mean Reciprocal Rank (MRR):
\begin{equation}
\mathrm{MRR} = \frac{1}{N} \sum_{i=1}^{N} \frac{1}{p_i}.
\end{equation}

Table~\ref{tab:typical} shows that ROG achieves the best performance across typical projection, intersection, union, and composite query types on both datasets.
In particular, ROG is consistently strong on deeper and more complex queries.
This suggests that decomposition and evidence grounding help mitigate the difficulty of long reasoning chains.

A clear trend is that the improvement is larger when the logical structure requires composing multiple operations (e.g., multi-hop projection or combined projection+set operators), where embedding-based approximations can accumulate error.
We also observe that the gains are generally more pronounced on NELL995, which is typically more sparse and thus benefits more from explicitly grounding each reasoning step in retrieved neighborhood evidence.

\subsection{Advantages of Chain-of-Thought Decomposition}
\label{sec:cot_advantage}

We analyze the impact of CoT-style decomposition by focusing on complex, operator-rich queries (Table~\ref{tab:complex}).
Embedding-based approaches such as BetaE address multi-hop reasoning by iteratively transforming latent representations, which can accumulate approximation errors as query depth and branching increase.
This is particularly problematic for negation, where a small error in the ``to-be-removed'' set can lead to a large change in the final answer set.

In contrast, ROG converts a complex query into a small sequence of simple sub-queries and explicitly materializes intermediate answer sets.
This turns multi-operator reasoning into a sequence of easier classification/filtering steps grounded in retrieved evidence.
Materialization also makes it possible to localize errors: if the final answer is wrong, we can inspect intermediate sets to find which sub-query failed.
Empirically, Table~\ref{tab:complex} shows consistent improvements (about 25\%--35\% MRR) on high-complexity negation queries.
These results support the hypothesis that decomposition plus retrieval provides a more stable pathway for complex logical inference than purely embedding-based set operations.

\section{Conclusion}

We presented ROG, a framework that integrates logical reasoning over knowledge graphs with large language model inference by combining query decomposition, query-aware neighborhood retrieval, and step-by-step reasoning.
ROG achieves consistent performance improvements over traditional embedding-based methods on standard KG benchmarks, with particularly strong gains on complex and negation-heavy query types.
Future work includes more efficient retrieval policies and tighter coupling between symbolic execution plans and prompting.

\bibliography{custom}

\end{document}